\documentclass[10pt,twocolumn,letterpaper]{article}

\usepackage{cvpr}
\usepackage{times}
\usepackage{epsfig}
\usepackage{graphicx}
\usepackage{amsmath}
\usepackage{amssymb}

\usepackage[pagebackref=true,breaklinks=true,letterpaper=true,colorlinks,bookmarks=false]{hyperref}
\usepackage{url}            
\usepackage{booktabs}       
\usepackage{threeparttable}
\usepackage{capt-of}
\usepackage{mathtools}
\usepackage{color}
\usepackage{wrapfig}
\usepackage{multirow}
\usepackage{subfigure}
\usepackage{appendix}
\usepackage{array}
\usepackage[linesnumbered,ruled,vlined]{algorithm2e}
\usepackage{caption}

\SetKwRepeat{Do}{do}{while}
\DeclareMathOperator*{\argmax}{arg\,max}

\cvprfinalcopy 

\def\cvprPaperID{2921} 

\ifcvprfinal\fi
\begin{document}
\renewcommand{\thefootnote}{\fnsymbol{footnote}}

\title{Densely Connected Search Space for More Flexible Neural Architecture Search}

\author{
Jiemin Fang$^{1}$\footnotemark[2] , Yuzhu Sun$^{1}$\footnotemark[2] , Qian Zhang$^{2}$, Yuan Li$^{2}$, Wenyu Liu$^{1}$, Xinggang Wang$^{1}$\footnotemark[3] \\
$^1$School of EIC, Huazhong University of Science and Technology $\; ^2$Horizon Robotics\\
\texttt{\{jaminfong, yzsun, liuwy, xgwang\}@hust.edu.cn}\\
\texttt{\{qian01.zhang, yuan.li\}@horizon.ai}
}

\maketitle

\begin{abstract}
    Neural architecture search (NAS) has dramatically advanced the development of neural network design. We revisit the search space design in most previous NAS methods and find the number and widths of blocks are set manually. However, block counts and block widths determine the network scale (depth and width) and make a great influence on both the accuracy and the model cost (FLOPs/latency). In this paper, we propose to search block counts and block widths by designing a densely connected search space, \ie, DenseNAS. The new search space is represented as a dense super network, which is built upon our designed routing blocks. In the super network, routing blocks are densely connected and we search for the best path between them to derive the final architecture. We further propose a chained cost estimation algorithm to approximate the model cost during the search. Both the accuracy and model cost are optimized in DenseNAS. For experiments on the MobileNetV2-based search space, DenseNAS achieves 75.3\% top-1 accuracy on ImageNet with only 361MB FLOPs and 17.9ms latency on a single TITAN-XP. The larger model searched by DenseNAS achieves 76.1\% accuracy with only 479M FLOPs. DenseNAS further promotes the ImageNet classification accuracies of ResNet-18, -34 and -50-B by 1.5\%, 0.5\% and 0.3\% with 200M, 600M and 680M FLOPs reduction respectively. The related code is available at \url{https://github.com/JaminFong/DenseNAS}.
\end{abstract}
\footnotetext[2]{The work is performed during the internship at Horizon Robotics.}
\footnotetext[3]{Corresponding author.}

\section{Introduction}
\label{sec: intro}
    In recent years, neural architecture search (NAS)~\cite{zoph2016neural, zoph2017learning, pham2018efficient, Real2018Regularized} has demonstrated great successes in designing neural architectures automatically and achieved remarkable performance gains in various tasks such as image classification~\cite{zoph2017learning, pham2018efficient}, semantic segmentation~\cite{DBLP:conf/nips/ChenCZPZSAS18, liu2019auto} and object detection~\cite{ghiasi2019fpn, Xu_2019_ICCV}. NAS has been a critically important topic for architecture designing.

\begin{figure}[t]
    \begin{center}
        \includegraphics[width=1.0\linewidth]{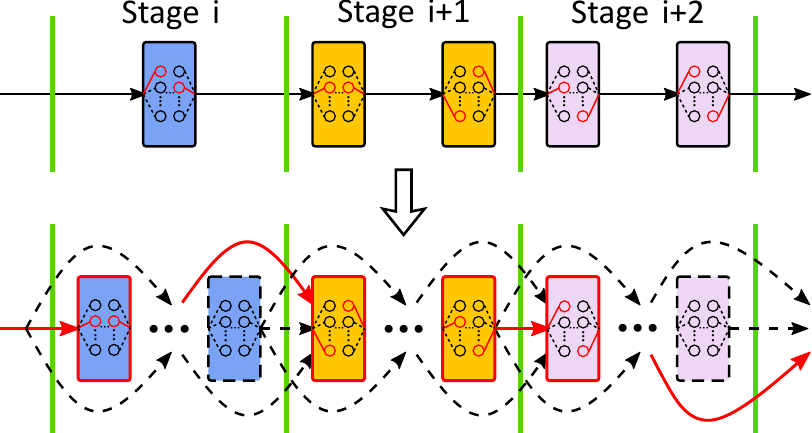}
    \end{center}
    \vspace{-8pt}
    \caption{Search space comparison between conventional methods and DenseNAS. 
    \emph{Upper}: Conventional search spaces manually set a fixed number of blocks in each stage. The block widths are set manually as well.
    \emph{Bottom}: The search space in DenseNAS allows more blocks with various widths in each stage. Each block is densely connected to its subsequent ones. We search for the best path (the red line) to derive the final architecture, in which the number of blocks in each stage and the widths of blocks are allocated automatically.
    }
    \label{fig: comp_ss}
    \vspace{-10pt}
\end{figure}

    In NAS research, the search space plays a crucial role that constrains the architectures in a prior-based set. The performance of architectures produced by NAS methods is strongly associated with the search space definition. A more flexible search space has the potential to bring in architectures with more novel structures and promoted performance. We revisit and analyze the search space design in most previous works~\cite{zoph2017learning, MnasNet, cai2018proxylessnas, fbnet}. For a clear illustration, we review the following definitions. \emph{Block} denotes a set of layers/operations in the network which output feature maps with the same spatial resolution and the same width (number of channels). \emph{Stage} denotes a set of sequential \emph{block}s whose outputs are under the same spatial resolution settings. Different \emph{block}s in the same stage are allowed to have various widths. Many recent works~\cite{cai2018proxylessnas, fbnet, chu2019fairnas} stack the inverted residual convolution modules (MBConv) defined in MobileNetV2~\cite{sandler2018mobilenetv2} to construct the search space. They search for different kernel sizes and expansion ratios in each MBConv. The depth is searched in terms of layer numbers in each block. The searched networks with MBConvs show high performance with low latency or few FLOPs. 
    
    In this paper, we aim to perform NAS in a more flexible search space. Our motivation and core idea are illustrated in Fig.~\ref{fig: comp_ss}. As the upper part of Fig.~\ref{fig: comp_ss} shows, the number of blocks in each stage and the width of each block are set manually and fixed during the search process. It means that the depth search is constrained within the block and the width search cannot be performed.
    It is worth noting that the scale (depth and width) setting is closely related to the performance of a network, which has been demonstrated in many previous theoretical studies~\cite{raghu2017expressive, lu2017expressive} and empirical results~\cite{gordon2018morphnet, tan2019efficientnet}. Inappropriate width or depth choices usually cause drastic accuracy degradation, significant computation cost, or unsatisfactory model latency. Moreover, we find that recent works~\cite{sandler2018mobilenetv2, cai2018proxylessnas, fbnet, chu2019fairnas} manually tune width settings to obtain better performance, which indicates the design of network width demands much prior-based knowledge and trial-and-error.
    
    We propose a densely connected search space to tackle the above obstacles and name our method as \emph{DenseNAS}. We show our novelly designed search space schematically in the bottom part of Fig.~\ref{fig: comp_ss}. Different from the search space design principles in the previous works~\cite{cai2018proxylessnas, fbnet}, we allow more blocks with various widths in one stage. Specifically, we design the \emph{routing block}s to construct the densely connected super network which is the representation of the search space. From the beginning to the end of the search space, the width of the routing block increases gradually to cover more width options. Every routing block is connected to several subsequent ones. This formulation brings in various paths in the search space and we search for the best path to derive the final architecture. As a consequence, the block widths and counts in each stage are allocated automatically. Our method extends the depth search into a more flexible space. Not only the number of layers within one block but also the number of blocks within one stage can be searched. The block width search is enabled as well. Moreover, the positions to conduct spatial down-sampling operations are determined along with the block counts search.

    We integrate our search space into the differentiable NAS framework by relaxing the search space. We assign a probability parameter to each output path of the routing block. During the search process, the distribution of probabilities is optimized. The final block connection paths in the super network are derived based on the probability distribution. To optimize the cost (FLOPs/latency) of the network, we design a \emph{chained estimation algorithm} targeted at approximating the cost of the model during the search.
    
    Our contributions can be summarized as follows.
    \begin{itemize}
        \item We propose a densely connected search space that enables network/block widths search and block counts search. It provides more room for searching better networks and further reduces expert designing efforts.
        \item We propose a chained cost estimation algorithm to precisely approximate the computation cost of the model during search, which makes the DenseNAS networks achieve high performance with low computation cost.
        \item In experiments, we demonstrate the effectiveness of our method by achieving SOTA performance on the MobileNetV2~\cite{sandler2018mobilenetv2}-based search space. Our searched network achieves 75.3\% accuracy on ImageNet~\cite{imagenet} with only 361MB FLOPs and 17.9ms latency on a single TITAN-XP.
        \item DenseNAS can further promote the ImageNet classification accuracies of ResNet-18, -34 and -50-B~\cite{he2016deep} by 1.5\%, 0.5\% and 0.3\% with 200M, 600M, 680M FLOPs and 1.5ms, 2.4ms, 6.1ms latency reduction respectively.
    \end{itemize}

\begin{figure*}[htbp]
    \vspace{-5pt}
    \begin{center}
        \includegraphics[width=1.0\linewidth]{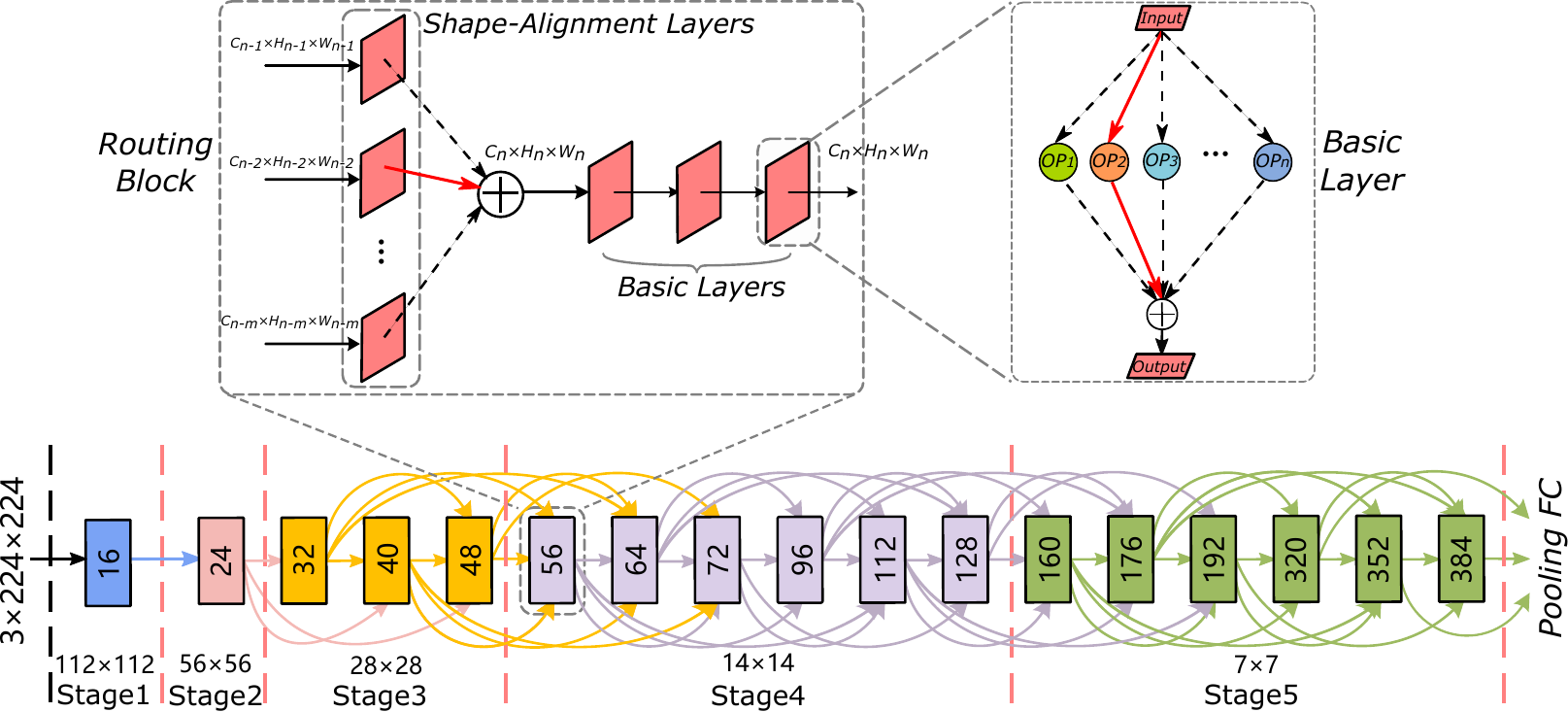}
    \end{center}
    \caption{We define our search space on three levels. \emph{Upper right}: A \textit{basic layer} that contains a set of candidate operations. \emph{Upper left}: The proposed \textit{routing block} which contains \textit{shape-alignment layer}s, an element-wise sum operation and some basic layers. It takes multiple input tensors and outputs one tensor. \emph{Bottom}: The proposed \textit{dense super network} which is constructed with densely connected routing blocks. Routing blocks in the same stage hold the same color. Architectures are searched within various path options in the super network.
    }
    \label{fig:searchspace}
    \vspace{-10pt}
\end{figure*}
\section{Related Work}

\paragraph{Search Space Design}
    NASNet~\cite{zoph2017learning} is the first work to propose a cell-based search space, where the cell is represented as a directed acyclic graph with several nodes inside. NASNet searches for the operation types and the topological connections in the cell and repeat the searched cell to form the whole network architecture. The depth of the architecture (\ie, the number of repetitions of the cell), the widths and the occurrences of down-sampling operations are all manually set. Afterwards, many works~\cite{liu2017progressive, pham2018efficient, Real2018Regularized, liu2018darts} adopt a similar cell-based search space. However, architectures generated by cell-based search spaces are not friendly in terms of latency or FLOPs. Then MnasNet~\cite{MnasNet} stacks MBConvs defined in MobileNetV2~\cite{sandler2018mobilenetv2} to construct a search space for searching efficient architectures. Some works~\cite{cai2018proxylessnas, eatnas, fbnet, ChamNet} simplify the search space by searching for the expansion ratios and kernel sizes of MBConv layers. 

    Some works study more about the search space. Liu \etal~\cite{liu2018hierarchical} proposes a hierarchical search space that allows flexible network topologies (directed acyclic graphs) at each level of the hierarchies. Auto-DeepLab~\cite{liu2019auto} creatively designs a two-level hierarchical search space for semantic segmentation networks. CAS~\cite{zhang2019customizable} customizes the search space design for real-time segmentation networks. RandWire~\cite{Xie_2019_ICCV} explores randomly wired architectures by designing network generators that produce new families of models for searching. Our proposed method designs a densely connected search space beyond conventional search constrains to generate the architecture with a better trade-off between accuracy and model cost.

\paragraph{NAS Method}
    Some early works~\cite{zoph2016neural, zoph2017learning, zhong2018practical} propose to search architectures based on reinforcement learning (RL) methods. Then evolutionary algorithm (EA) based methods~\cite{dong2018dpp, liu2018hierarchical, Real2018Regularized} achieve great performance. However, RL and EA based methods bear huge computation cost. As a result, ENAS~\cite{pham2018efficient} proposes to use weight sharing for reducing the search cost.

    Recently, the emergence of differentiable NAS methods~\cite{liu2018darts, cai2018proxylessnas, fbnet} and one-shot methods~\cite{brock2017smash, Understanding} greatly reduces the search cost and achieves superior results. DARTS~\cite{liu2018darts} is the first work to utilize the gradient-based method to search neural architectures. They relax the architecture representation as a super network by assigning continuous weights to the candidate operations. They first search on a small dataset, \eg, CIFAR-10~\cite{krizhevsky2009learning}, and then apply the architecture to a large dataset, \eg, ImageNet~\cite{DBLP:conf/cvpr/DengDSLL009}, with some manual adjustments. ProxylessNAS~\cite{cai2018proxylessnas} reduces the memory consumption by adopting a dropping path strategy and conducts search directly on the large scale dataset, \ie, ImageNet. FBNet~\cite{fbnet} searches on the subset of ImageNet and uses the Gumbel Softmax function~\cite{JangGP17, MaddisonMT17} to better optimize the distribution of architecture probabilities. TAS~\cite{dong2019network} utilizes a differentiable NAS scheme to search and prune the width and depth of the network and uses knowledge distillation (KD)~\cite{hinton2015distilling} to promote the performance of the pruned network. FNA~\cite{Fang*2020Fast} proposes to adapt the neural network to new tasks with low cost by a parameter remapping mechanism and differentiable NAS. It is challenging for differentiable/one-shot NAS methods to search for more flexible architectures as they need to integrate all sub-architectures into the super network. The proposed DenseNAS tends to solve this problem by integrating a densely connected search space into the differentiable paradigm and explores more flexible search schemes in the network.

\section{Method}
    In this section, we first introduce how to design the search space targeted at a more flexible search. A \emph{routing block} is proposed to construct the densely connected super network. Secondly, we describe the method of relaxing the search space into a continuous representation. Then, we propose a \emph{chained cost estimation} algorithm to approximate the model cost during the search. Finally, we describe the whole search procedure.
 
    \subsection{Densely Connected Search Space}
    As shown in Fig.~\ref{fig:searchspace}, we define our search space using the following three terms, \ie, (\emph{basic layer}, \emph{routing block} and \emph{dense super Network}). Firstly, a \emph{basic layer} is defined as a set of all the candidate operations. Then we propose a novel \emph{routing block} which can aggregate tensors from different routing blocks and transmit tensors to multiple other routing blocks. Finally, the search space is constructed as a \emph{dense super network} with many routing blocks where there are various paths to transmit tensors.
   
    \subsubsection{Basic Layer}
    \label{sssec:layer}
    We define the \emph{basic layer} to be the elementary structure in our search space. One basic layer represents a set of candidate operations which include MBConvs and the skip connection. MBConvs are with kernel sizes of $\{3, 5, 7\}$ and expansion ratios of $\{3, 6\}$. The skip connection is for the depth search. If the skip connection is chosen, the corresponding layer is removed from the resulting architecture.

    \subsubsection{Routing Block}
    For the purpose of establishing various paths in the super network, we propose the \emph{routing block} with the ability of aggregating tensors from preceding routing blocks and transmit tensors to subsequent ones. We divide the routing block into two parts, \emph{shape-alignment layers} and \emph{basic layers}.

    Shape-alignment layers exist in the form of several parallel branches, while every branch is a set of candidate operations. They take input tensors with different shapes (including widths and spatial resolutions) which come from multiple preceding routing blocks and transform them into tensors with the same shape. As shape-alignment layers are required for all routing blocks, we exclude the skip connection in candidate operations of them. Then tensors processed by shape-alignment layers are aggregated and sent to several basic layers. The subsequent basic layers are used for feature extraction whose depth can also be searched.

    \subsubsection{Dense Super Network}
    \label{sssec: network}
    Many previous works~\cite{MnasNet, cai2018proxylessnas, fbnet} manually set a fixed number of blocks, and retain all the blocks for the final architecture. Benefiting from the aforementioned structures of routing blocks, we introduce more routing blocks with various widths to construct the \emph{dense super network} which is the representation of the search space. The final searched architecture is allowed to select a subset of the routing blocks and discard the others, giving the search algorithm more room.
    
    We define the super network as $\mathcal{N}_{sup}$ and assume it to consist of $N$ routing blocks, $\mathcal{N}_{sup} = \{B_1, B_2, ..., B_N\}$. The network structure is shown in Fig.~\ref{fig:searchspace}. We partition the entire network into several stages. As Sec.~\ref{sec: intro} defines, each stage contains routing blocks with various widths and the same spatial resolution. From the beginning to the end of the super network, the widths of routing blocks grow gradually. In the early stage of the network, we set a small growing stride for the width because large width settings in the early network stage will cause huge computational cost. The growing stride becomes larger in the later stages. This design principle of the super network allows more possibilities of block counts and block widths.

    We assume that each routing block in the super network connects to $M$ subsequent ones. We define the connection between the routing block $B_i$ and its subsequent routing block $B_j$ ($j > i$) as $C_{ij}$. The spatial resolutions of $B_i$ and $B_j$ are $H_i \times W_i$ and $H_j \times W_j$ respectively (normally $H_i = W_i$ and $H_j = W_j$). We set some constraints on the connections to avoid the stride of the spatial down-sampling exceeding $2$. Specifically, $C_{ij}$ only exists when $j - i \leq M$ and $H_i / H_j \leq 2$. Following the above paradigms, the search space is constructed as a dense super network based on the connected routing blocks. 
    
    \subsection{Relaxation of Search Space}
    \label{ssec: relax}
    We integrate our search space by relaxing the architectures into continuous representations. The relaxation is implemented on both the basic layer and the routing block. We can search for architectures via back-propagation in the relaxed search space.

    \subsubsection{Relaxation in the Basic Layer}
    Let $\mathcal{O}$ be the set of candidate operations described in Sec.~\ref{sssec:layer}. We assign an architecture parameter $\alpha_o^\ell$ to the candidate operation $o \in \mathcal{O}$ in basic layer $\ell$. We relax the basic layer by defining it as a weighted sum of outputs from all candidate operations. The architecture weight of the operation is computed as a \emph{softmax} of architecture parameters over all operations in the basic layer:
    \begin{equation}
        w_o^\ell = \frac{\exp(\alpha_o^\ell)}{\sum_{o' \in \mathcal{O}} \exp(\alpha_{o'}^\ell)}
        \label{eq: op_weight}.
    \end{equation}
    The output of basic layer $\ell$ can be expressed as 
    \begin{equation}
        x_{\ell+1} = \sum_{o \in \mathcal{O}} w_o^\ell \cdot o(x_\ell),
    \end{equation}
    where $x_\ell$ denotes the input tensor of basic layer $\ell$.

    \subsubsection{Relaxation in the Routing Block}
    We assume that the routing block $B_i$ outputs the tensor $b_i$ and connects to $m$ subsequent blocks. To relax the block connections as a continuous representation, we assign each output path of the block an architecture parameter. Namely the path from $B_i$ to $B_j$ has a parameter $\beta_{ij}$. Similar to how we compute the architecture weight of each operation above, we compute the probability of each path using a \emph{softmax} function over all paths between the two routing blocks:
    \begin{equation}
        p_{ij} = \frac{\exp(\beta_{ij})}{\sum_{k=1}^m \exp(\beta_{ik})}.
    \end{equation}
    For routing block $B_i$, we assume it takes input tensors from its $m'$ preceding routing blocks ($B_{i-m'}$, $B_{i-m'+1}$, $B_{i-m'+2}$ ... $B_{i-1}$). As shown in Fig.~\ref{fig:searchspace}, the input tensors from these routing blocks differ in terms of width and spatial resolution. Each input tensor is transformed to a same size by the corresponding branch of shape-alignment layers in $B_i$. Let $H_{ik}$ denotes the $k$th transformation branch in $B_i$ which is applied to the input tensor from $B_{i-k}$, where $k=1 \dots m'$. Then the input tensors processed by shape-alignment layers are aggregated by a weighted-sum using the path probabilities,
    \begin{equation}
        x_i = \sum_{k=1}^{m'} p_{i-k, i} \cdot H_{ik}(x_{i-k}).
    \end{equation}
    It is worth noting that the path probabilities are normalized on the output dimension but applied on the input dimension (more specifically on the branches of shape-alignment layers). One of the shape-alignment layers is essentially a weighted-sum mixture of the candidate operations. The layer-level parameters $\alpha$ control which operation to be selected, while the outer block-level parameters $\beta$ determine how blocks connect.
 
    \subsection{Chained Cost Estimation Algorithm}
    \label{sssec: chain est}
    We propose to optimize both the accuracy and the cost (latency/FLOPs) of the model. To this end, the model cost needs to be estimated during the search. In conventional cascaded search spaces, the total cost of the whole network can be computed as a sum of all the blocks. Instead, the global effects of connections on the predicted cost need to be taken into consideration in our densely connected search space. We propose a \emph{chained cost estimation} algorithm to better approximate the model cost.

    We create a lookup table which records the cost of each operation in the search space. The cost of every operation is measured separately. During the search, the cost of one basic layer is estimated as follows,
    \begin{equation}
        \mathtt{cost}^\ell = \sum_{o \in \mathcal{O}} w_o^\ell \cdot \mathtt{cost}_o^\ell,
    \end{equation}
    where $\mathtt{cost}_o^\ell$ refers to the pre-measured cost of operation $o \in \mathcal{O}$ in layer $\ell$. We assume there are $N$ routing blocks in total ($B_1, \dots, B_N$). To estimate the total cost of the whole network in the densely connected search space, we define the chained cost estimation algorithm as follows.
    \begin{equation}
    \label{eq: chain cost}
        \begin{aligned}
            \tilde{\mathtt{cost}}^N =&\  \mathtt{cost}^N_{b} \\
            \tilde{\mathtt{cost}}^i =&\  \mathtt{cost}^i_{b} + \sum_{j=i+1}^{i+m} p_{ij} \cdot (\mathtt{cost}^{ij}_{align} + \mathtt{cost}^j_{b}),
        \end{aligned}
    \end{equation}
    where $\mathtt{cost}^i_{b}$ denotes the total cost of all the basic layers of $B_i$ which can be computed as a sum $\mathtt{cost}_b^i = \sum_\ell \mathtt{cost}_b^{i, \ell}$, $m$ denotes the number of subsequent routing blocks to which $B_i$ connects, $p_{ij}$ denotes the path probability between $B_i$ and $B_j$, and $\mathtt{cost}_{align}^{ij}$ denotes the cost of the shape-alignment layer in block $B_j$ which processes the data from block $B_i$.

    The cost of the whole architecture can thus be obtained by computing $\tilde{\mathtt{cost}}^1$ with a recursion mechanism,
    \begin{equation}
        \mathtt{cost} = \tilde{\mathtt{cost}}^1.
    \end{equation}
    We design a loss function with the cost-based regularization to achieve the multi-objective optimization:
    \begin{equation}
        \label{eq: loss}
        \mathcal{L}(w, \alpha, \beta) = \mathcal{L}_{CE} + \lambda \log_\tau\mathtt{cost},
    \end{equation}
    where $\lambda$ and $\tau$ are the hyper-parameters to control the magnitude of the model cost term.
    
    \subsection{Search Procedure}
    \label{sec: search_pro}
    Benefiting from the continuously relaxed representation of the search space, we can search for the architecture by updating the architecture parameters (introduced in Sec.~\ref{ssec: relax}) using stochastic gradient descent. We find that at the beginning of the search process, all the weights of the operations are under-trained. The operations or architectures which converge faster are more likely to be strengthened, which leads to shallow architectures. To tackle this, we split our search procedure into two stages. In the first stage, we only optimize the weights for enough epochs to get operations sufficiently trained until the accuracy of the model is not too low. In the second stage, we activate the architecture optimization. We alternatively optimize the operation weights by descending $\nabla_w \mathcal{L}_{train}(w, \alpha, \beta)$ on the training set, and optimize the architecture parameters by descending $\nabla_{\alpha, \beta} \mathcal{L}_{val}(w, \alpha, \beta)$ on the validation set. Moreover, a dropping-path training strategy~\cite{Understanding,cai2018proxylessnas} is adopted to decrease memory consumption and decouple different architectures in the super network.

    When the search procedure terminates, we derive the final architecture based on the architecture parameters $\alpha, \beta$. At the layer level, we select the candidate operation with the maximum architecture weight, \ie, $\argmax_{o \in \mathcal{O}} \alpha_o^\ell$. At the network level, we use the Viterbi algorithm~\cite{forney1973viterbi} to derive the paths connecting the blocks with the highest total transition probability based on the output path probabilities. Every block in the final architecture only connects to the next one.

\section{Experiments}
    In this section, we first show the performance with the MobileNetV2~\cite{sandler2018mobilenetv2}-based search space on ImageNet~\cite{imagenet} classification. Then we apply the architectures searched on ImageNet to object detection on COCO~\cite{COCO}. We further extend our DenseNAS to the ResNet~\cite{he2016deep}-based search space. Finally, we conduct some ablation studies and analysis. The implementation details are provided in the appendix.

\begin{table}[thbp]
    \centering
    \caption{Our results on the ImageNet classification with the MobileNetV2-based search space compared with other methods. Our models achieve higher accuracies with lower latencies. For GPU latency, we measure all the models with the same setup (on one TITAN-XP with a batch size of 32).} 
    \label{tab: mb_results}
    \begin{threeparttable}
    \resizebox{1\columnwidth}{!}{
        \begin{tabular}{lp{0.1\columnwidth}p{0.15\columnwidth}p{0.15\columnwidth}p{0.23\columnwidth}}
        \toprule
        \multirow{2}*{\textbf{Model}} & \multirow{2}*{\textbf{FLOPs}} & \textbf{GPU Latency} & \textbf{Top-1 Acc(\%)} & \textbf{Search Time (GPU hours)}\\ 
        \midrule
        1.4-MobileNetV2~\cite{sandler2018mobilenetv2} & 585M & 28.0ms & 74.7 & - \\
        NASNet-A~\cite{zoph2017learning} & 564M & - & 74.0 & 48K \\ 
        AmoebaNet-A~\cite{Real2018Regularized} & 555M & - & 74.5 & 76K \\
        DARTS~\cite{liu2018darts} & 574M & 36.0ms & 73.3 & 96 \\
        RandWire-WS~\cite{Xie_2019_ICCV} & 583M & - & 74.7 & - \\
        DenseNAS-Large & \textbf{479M} & 28.9ms & \textbf{76.1} & 64\\
        \midrule
        1.0-MobileNetV1~\cite{howard2017mobilenets} & 575M & 16.8ms & 70.6 & - \\
        1.0-MobileNetV2~\cite{sandler2018mobilenetv2} & 300M & 19.5ms & 72.0 & - \\
        FBNet-A~\cite{fbnet} & 249M & 15.8ms & 73.0 & 216 \\
        DenseNAS-A & 251M & 13.6ms & 73.1 & 64 \\
        \midrule
        MnasNet~\cite{MnasNet} & 317M & 19.7ms & 74.0 & 91K \\ 
        FBNet-B~\cite{fbnet} & 295M & 18.9ms & 74.1 & 216 \\
        Proxyless(mobile)~\cite{cai2018proxylessnas} & 320M & 21.3ms & 74.6 & 200 \\
        DenseNAS-B & \textbf{314M} & \textbf{15.4ms} & \textbf{74.6} & 64 \\
        \midrule
        MnasNet-92~\cite{MnasNet} & 388M & - & 74.8 & 91K \\
        FBNet-C~\cite{fbnet} & 375M & 22.1ms & 74.9 & 216 \\
        Proxyless(GPU)~\cite{cai2018proxylessnas} & 465M & 22.1ms & 75.1 & 200 \\
        DenseNAS-C & \textbf{361M} & \textbf{17.9ms} & \textbf{75.3} & 64 \\
        \midrule
        Random Search & 360M & 26.9ms & 74.3 & 64 \\
        \bottomrule
        \end{tabular}}
        \end{threeparttable}
    \vspace{-10pt}
    \end{table}

    \subsection{Performance on MobileNetV2-based Search Space}
\begin{figure}[thbp]
    \centering
    \includegraphics[width=0.8\linewidth]{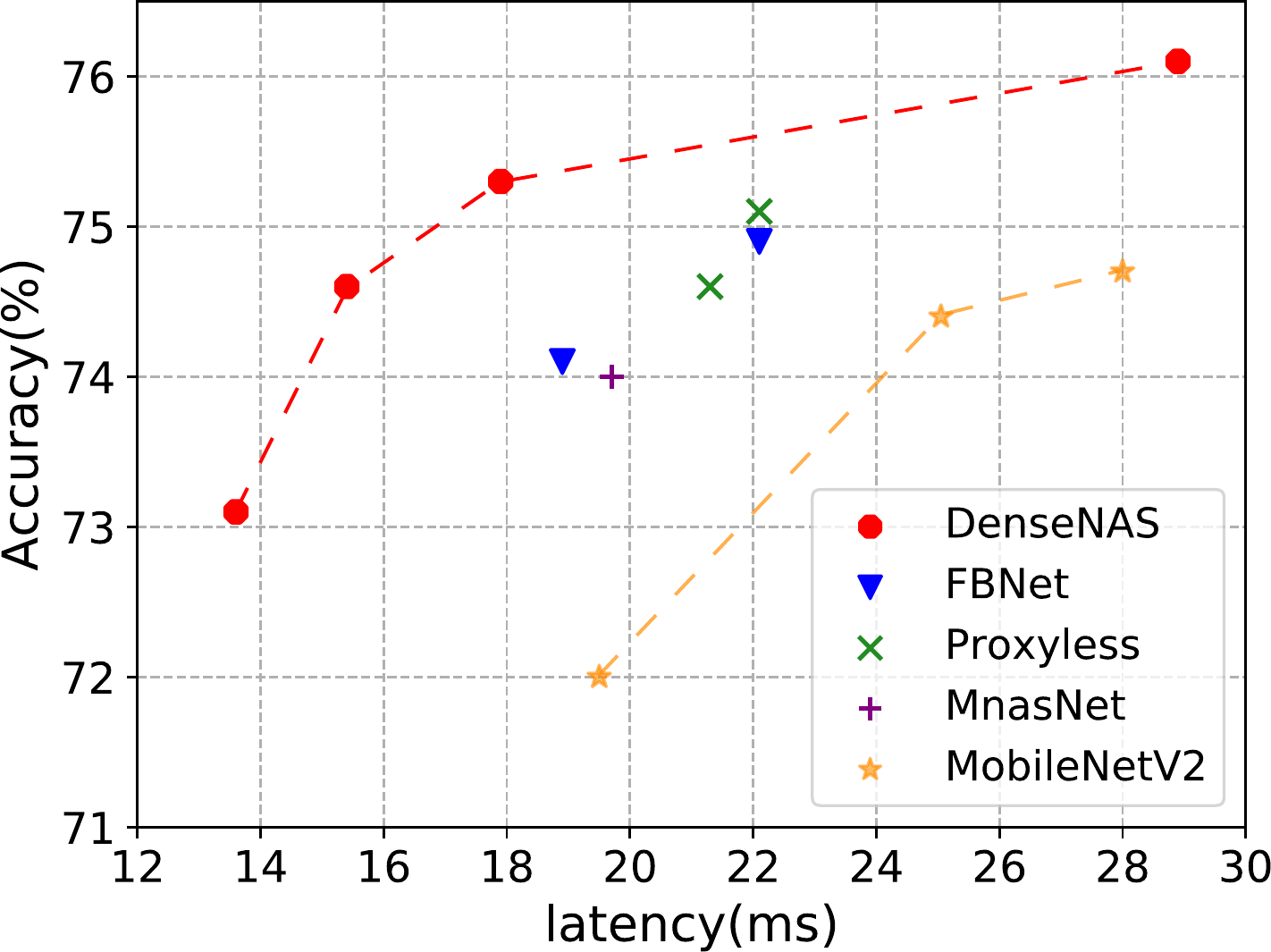}
    \caption{The comparison of model performance on ImageNet under the MobileNetV2-based search spaces.}
    \label{fig: comp_mobile}
    \vspace{-8pt}
\end{figure}

\begin{figure}[htbp]
    \centering
    \includegraphics[width=1\columnwidth]{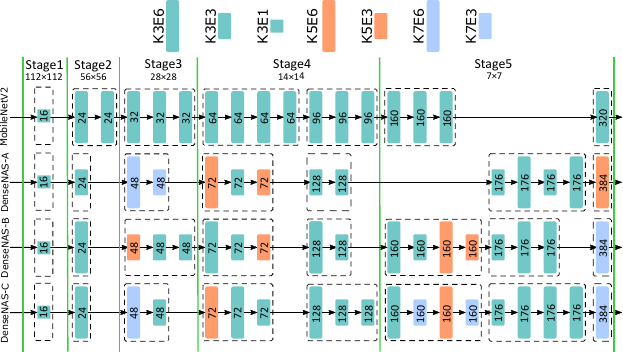}
    \vspace{-15pt}
    \caption{Visualization of the searched architectures. We use rectangles with different colors and widths to denote the layer operations. $KxEy$ denotes the MBConv with kerner size $x \times x$ and expansion ratio $y$. We label the width in each layer. Layers in the same block are contained in the dashed box. We separate the stages with the green lines.}
    \label{fig: architecture}
    \vspace{-12pt}
\end{figure}

    We implement DenseNAS on the MobileNetV2~\cite{sandler2018mobilenetv2}-based search space, set the GPU latency as our secondary optimization objective, and search models with different sizes under multiple latency optimization magnitudes (defined in Eq.~\ref{eq: loss}). The ImageNet results are shown in Tab.~\ref{tab: mb_results}. We divide Tab.~\ref{tab: mb_results} into several parts and compare DenseNAS models with both manually designed models~\cite{howard2017mobilenets, sandler2018mobilenetv2} and NAS models. DenseNAS achieves higher accuracies with both fewer FLOPs and lower latencies. Note that for FBNet-A, the group convolution in the 1$\times$1 conv and the channel shuffle operation are used, which do not exist in FBNet-B, -C and Proxyless. In the compared NAS methods~\cite{MnasNet,cai2018proxylessnas,fbnet}, the block counts and block widths in the search space are set and adjusted manually. DenseNAS allocates block counts and block widths automatically. We further visualize the results in Fig.~\ref{fig: comp_mobile}, which clearly demonstrates that DenseNAS achieves a better trade-off between accuracy and latency. The searched architectures are shown in Fig.~\ref{fig: architecture}.

    \subsection{Generalization Ability on COCO Object Detection}
    We apply the searched DenseNAS networks on the COCO~\cite{COCO} object detection task to evaluate the generalization ability of DenseNAS networks and show the results in Tab.~\ref{tab: det}. We choose two commonly used object detection frameworks RetinaNet~\cite{lin2017focal} and SSDLite~\cite{liu2016ssd,sandler2018mobilenetv2} to conduct our experiments. All the architectures shown in Tab.~\ref{tab: det} are utilized as the backbone networks in the detection frameworks. The experiments are performed based on the MMDetection~\cite{chen2019mmdetection} framework. 

    We compare our results with both manually designed and NAS models. Results of MobileNetV2~\cite{sandler2018mobilenetv2}, FBNet~\cite{fbnet} and ProxylessNAS~\cite{cai2018proxylessnas} are obtained by our re-implementation and all models are trained under the same settings and hyper-parameters for fair comparisons. DetNAS~\cite{DBLP:journals/corr/abs-1903-10979} is a recent work that aims at searching the backbone architectures directly on object detection. Though DenseNAS searches on the ImageNet classification task and applies the searched architectures on detection tasks, our DenseNAS models still obtain superior detection performance in terms of both accuracy and FLOPs. The superiority over the compared methods demonstrates the great generalization ability of DenseNAS networks.

\begin{table}[tbp]
    \centering
    \caption{Object detection results on COCO. The FLOPs are calculated with $1088 \times 800$ input.}
    \label{tab: det}
    \begin{threeparttable}
    \resizebox{1\columnwidth}{!}{
        \begin{tabular}{l | c | c | c | c }
        \toprule
        \multicolumn{2}{l|}{\textbf{Method}} & \textbf{Params} & \textbf{FLOPs} & \textbf{mAP(\%)} \\
        \hline
        MobileNetV2~\cite{sandler2018mobilenetv2} & \multirow{6}*{RetinaNet} & 11.49M & 133.05B & 32.8\\
        DenseNAS-B && 12.69M & 133.09B & \textbf{34.3}\\
        \cline{1-1} \cline{3-5}
        DetNAS~\cite{DBLP:journals/corr/abs-1903-10979} && 13.41M & 133.26B & 33.3\\
        FBNet-C~\cite{fbnet} && 12.65M & 134.17B & 34.9\\
        Proxyless(GPU)~\cite{cai2018proxylessnas} && 14.62M & 135.81B & 35\\
        DenseNAS-C && 13.24M & 133.91B & \textbf{35.1}\\
        \hline
        MobileNetV2~\cite{sandler2018mobilenetv2} &\multirow{5}*{SSDLite}&4.3M &0.8B &22.1 \\
        Mnasnet-92~\cite{MnasNet} & &5.3M &1.0B &22.9 \\
        FBNet-C~\cite{fbnet} && 6.27M & 1.06B & 22.9\\
        Proxyless(GPU)~\cite{cai2018proxylessnas} && 7.91M & 1.24B & 22.8\\
        DenseNAS-C && 6.87M & 1.05B & \textbf{23.1}\\
        \bottomrule
    \end{tabular}}
    \end{threeparttable}
    \vspace{-7pt}
\end{table}

\begin{table}[htbp]
    \centering
    \caption{ImageNet classification results of ResNets and DenseNAS networks searched on the ResNet-based search spaces.} 
    \label{tab: resnet_results}
    \begin{threeparttable}
    \resizebox{1\columnwidth}{!}{
        \begin{tabular}{lp{0.15\columnwidth}p{0.15\columnwidth}p{0.15\columnwidth}p{0.15\columnwidth}}
        \toprule
        \multirow{2}*{\textbf{Model}} & \multirow{2}*{\textbf{Params}} & \multirow{2}*{\textbf{FLOPs}} & \textbf{GPU Latency} & \textbf{Top-1 Acc(\%)}\\
        \midrule
        ResNet-18~\cite{he2016deep} & 11.7M & 1.81B & 13.5ms & 72.0 \\
        DenseNAS-R1 & 11.1M & 1.61B & 12.0ms & 73.5 \\
        \midrule
        ResNet-34~\cite{he2016deep} & 21.8M & 3.66B & 24.6ms & 75.3 \\
        DenseNAS-R2 & 19.5M & 3.06B & 22.2ms & 75.8 \\ 
        \midrule
        ResNet-50-B~\cite{he2016deep} & 25.6M & 4.09B & 47.8ms & 77.7 \\
        RandWire-WS, C=109~\cite{Xie_2019_ICCV} & 31.9M & 4.0B & - & 79.0 \\
        DenseNAS-R3 & 24.7M & 3.41B & 41.7ms & 78.0 \\ 
        \bottomrule
        \end{tabular}}
        \end{threeparttable}
        \vspace{-10pt}
\end{table}

    \subsection{Performance on ResNet-based Search Space}
    We apply our DenseNAS framework on the ResNet~\cite{he2016deep}-based search space to further evaluate the generalization ability of our method. It is convenient to implement DenseNAS on ResNet~\cite{he2016deep} as we set the candidate operations in the basic layer as the basic block defined in ResNet~\cite{he2016deep} and the skip connection. The ResNet-based search space is also constructed as a densely connected super network.

\begin{figure}[t!]
    \centering
    \includegraphics[width=0.8\linewidth]{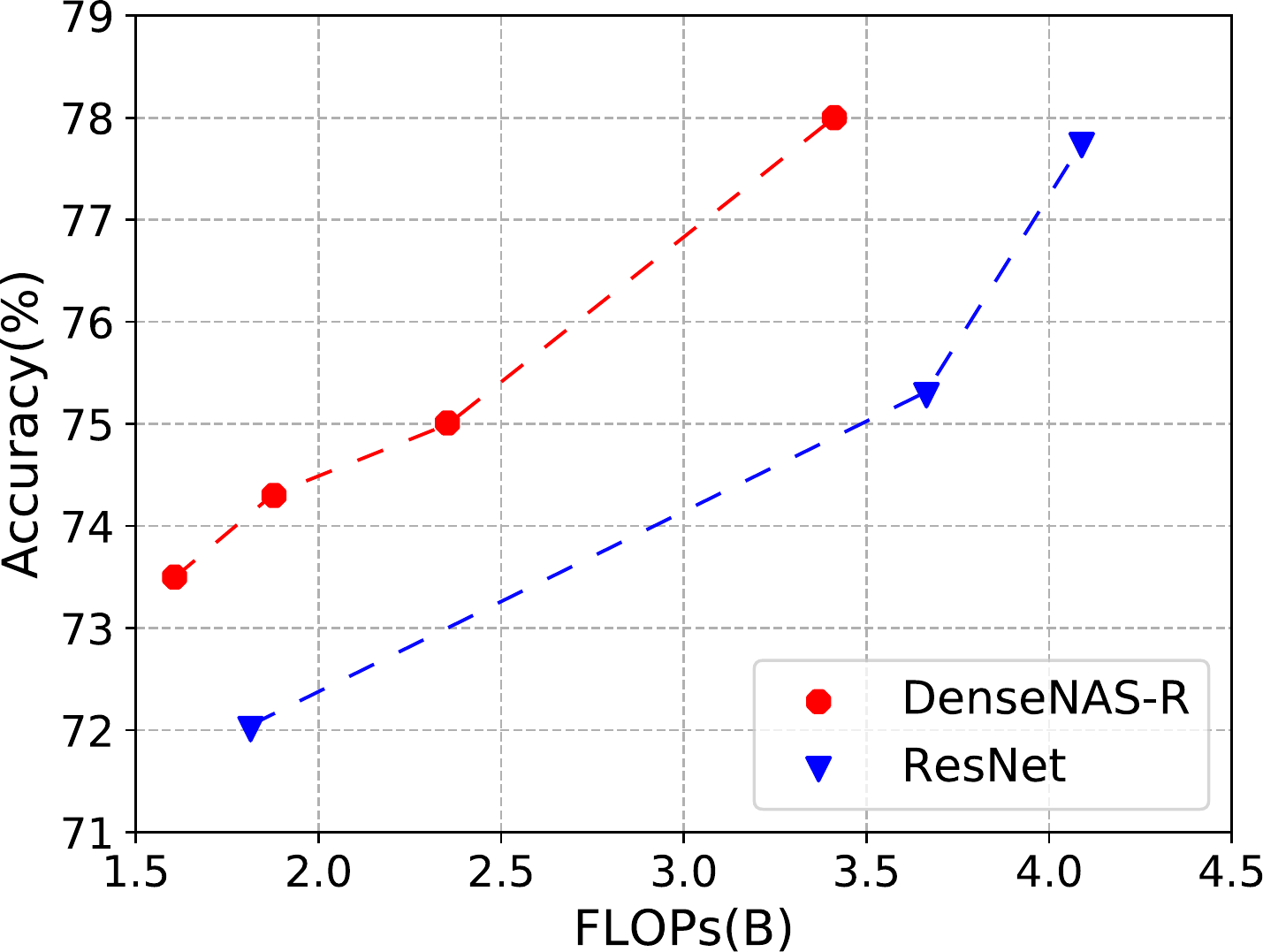}
    \caption{Graphical comparisons between ResNets and DenseNAS networks on the ResNet-based search space.}
    \label{fig: comp_res}
    \vspace{-10pt}
\end{figure}

    We search for several architectures with different FLOPs and compare them with the original ResNet models on the ImageNet~\cite{imagenet} classification task in Tab.~\ref{tab: resnet_results}. We further replace all the basic blocks in DenseNAS-R2 with the bottleneck blocks and obtain DenseNAS-R3 to compare with ResNet-50-B and the NAS model RandWire-WS, C=109~\cite{Xie_2019_ICCV} (WS, C=109). Though WS, C=109 achieves a higher accuracy, the FLOPs increases ~600M, which is a great number, 17.6\% of DenseNAS-R3. Besides, WS C=109 uses separable convolutions which greatly decrease the FLOPs while DenseNAS-R3 only contains plain convolutions. Moreover, RandWire networks are unfriendly to inference on existing hardware for the complicated connection patterns. Our proposed DenseNAS promotes the accuracy of ResNet-18, -34 and -50-B by 1.5\%, 0.5\% and 0.3\% with 200M, 600M, 680M fewer FLOPs and 1.5ms, 2.4ms, 6.1ms lower latency respectively. We visualize the comparison results in Fig.~\ref{fig: comp_res} and the performance on the ResNet-based search space further demonstrates the great generalization ability and effectiveness of DenseNAS.

    \subsection{Ablation Study and Analysis}
    \paragraph{Comparison with Other Search Spaces}
    \label{sssec: fixed-block}
    To further demonstrate the effectiveness of our proposed densely connected search space, we conduct the same search algorithm used in DenseNAS on the search spaces of FBNet and ProxylessNAS as well as a new search space which is constructed following the settings of block counts and block widths in MobileNetV2. The three search spaces are denoted as FBNet-SS, Proxyless-SS and MBV2-SS respectively. All the search/training settings and hyper-parameters are the same as that we use for DenseNAS. The results are shown in Tab.~\ref{tab: fixed-block} and DenseNAS achieves the highest accuracy with the lowest latency.

\begin{table}[htbp]
    \centering
    \caption{Comparisons with other search spaces on ImageNet. SS: Search Space. MBV2: MobileNetV2.} 
    \begin{threeparttable}
    \resizebox{1\columnwidth}{!}{
        \begin{tabular}{lcccc}
        \toprule
        \textbf{Search Space} & \textbf{FLOPs} & \textbf{GPU Latency} & \textbf{Top-1 Acc(\%)}\\
        \midrule
        FBNet-SS~\cite{fbnet} & 369M & 25.6ms & 74.9 \\
        Proxyless-SS~\cite{cai2018proxylessnas} & 398M & 18.9ms & 74.8 \\
        MBV2-SS~\cite{sandler2018mobilenetv2} & 383M & 32.1ms & 74.7 \\
        \midrule
        DenseNAS & 361M & 17.9ms & 75.3\\
        \bottomrule
        \end{tabular}}
        \end{threeparttable}
    \vspace{-15pt}
\label{tab: fixed-block}
\end{table}

    \vspace{-5pt}
    \paragraph{Comparison with Random Search}
    As random search~\cite{li2019random, sciuto2019evaluating} is treated as an important baseline to validate NAS methods. We conduct random search experiments and show the results in Tab.~\ref{tab: mb_results}. We randomly sample 15 models in our search space whose FLOPs are similar to DenseNAS-C. Then we train every model for 5 epochs on ImageNet. Finally, we select the one with the highest validation accuracy and train it under the same settings as DenseNAS. The total search cost of the random search is the same as DenseNAS. We observe that DenseNAS-C is 1\% accuracy higher compared with the randomly searched model, which proves the effectiveness of DenseNAS.

\begin{table}[t]
    \centering
    \caption{Comparison with different connection number settings on ImageNet.} 
    \begin{threeparttable}
    \resizebox{1\columnwidth}{!}{
        \begin{tabular}{cccc}
        \toprule
        \textbf{Connect Num} & \textbf{GPU Latency} & \textbf{Top-1 Acc(\%)} & \textbf{Search Epochs} \\
        \midrule
        3 & 18.9ms & 74.8 & 150 \\
        4 & 17.9ms & 75.3 & 150 \\
        5 & 19.2ms & 74.6 & 150 \\
        5 & 16.7ms & 74.9 & 200 \\
        \bottomrule
        \end{tabular}}
        \end{threeparttable}
        \vspace{-10pt}
\label{tab: connect}
\end{table}
    
    \vspace{-10pt}
    \paragraph{The Number of Block Connections}
    We explore the effect of the maximum number of connections between routing blocks in the search space. We set the maximum connection number as 4 in DenseNAS. Then we try more options and show the results in Tab.~\ref{tab: connect}. When we set the connection number to 3, the searched model gets worse performance. We attribute this to the search space shrinkage which causes the loss of many possible architectures with good performance. As we set the number to 5 and the search process takes the same number of epochs as DenseNAS, \ie 150 epochs. The performance of the searched model is not good, even worse than that of the connection number 3. Then we increase the search epochs to 200 and the search process achieves a comparable result with DenseNAS. This phenomenon indicates that larger search spaces need more search cost to achieve comparable/better results with/than smaller search spaces with some added constraints.

\begin{figure}[t]
    \vspace{-15pt}
    \centering
    \includegraphics[width=1\columnwidth]{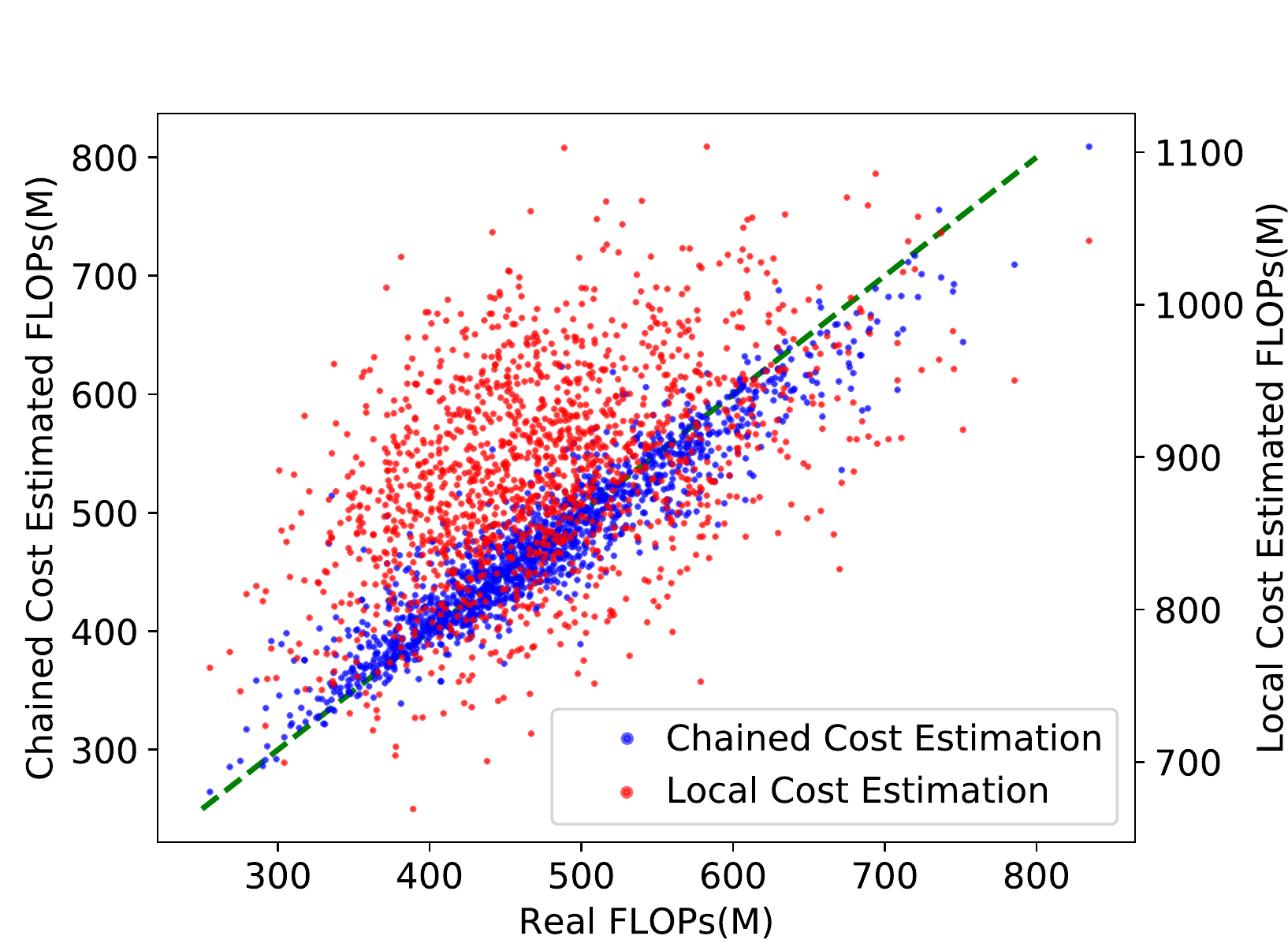}
    \caption{Predicted values of FLOPs computed by chained cost estimation and local cost estimation algorithm.}
    \label{fig: comp_est}
    \vspace{-15pt}
\end{figure}

    \vspace{-10pt}
    \paragraph{Cost Estimation Method}
    \label{sec: cost-est-exp}
    As the super network is densely connected and the final architecture is derived based on the total transition probability, the model cost estimation needs to take the effects of all the path probabilities on the whole network into consideration. We try a \emph{local cost estimation} strategy that does not involve the global connection effects on the whole super network. Specifically, we compute the cost of the whole network by summing the cost of every routing block during the search as follows, while the transition probability $p_{ji}$ is only used for computing the cost of each individual block rather than the whole network.
    \begin{equation}
        \mathtt{cost} = \sum_i^B (\sum_{j=i-m}^{j=i-1} p_{ji} \cdot \mathtt{cost}_{align}^{ji} + \mathtt{cost}_b^i),
    \end{equation}
    where all definitions in the equation are the same as that in Eq.~\ref{eq: chain cost}. We randomly generate the architecture parameters ($\alpha$ and $\beta$) to derive the architectures. Then we draw the approximated cost values computed by \emph{local cost estimation} and our proposed \emph{chained cost estimation} respectively, and compare with the real cost values in Fig.~\ref{fig: comp_est}. In this experiment, we take FLOPs as the model cost because the FLOPs is easier to measure than latency. 1,500 models are sampled in total. The results show that the predicted cost values computed by our \emph{chained cost estimation} algorithm has a much stronger correlation with the real values and approximate more to the real ones. As the predicted values are computed based on the randomly generated architecture parameters which are not binary parameters, there are still differences between the predicted and real values.

    \vspace{-10pt}
    \paragraph{Architecture Analysis}
    We visualize the searched architectures in Fig.~\ref{fig: architecture}. It shows that DenseNAS-B and -C have one more block in the last stage than other architectures, which indicates enlarging the depth in the last stage of the network tends to obtain a better accuracy. Moreover, the smallest architecture DenseNAS-A whose FLOPs is only 251M has one fewer block than DenseNAS-B and -C to decrease the model cost. The structures of the final searched architectures show the great flexibility of DenseNAS.

\section{Conclusion}
    We propose a densely connected search space for more flexible architecture search, DenseNAS. We tackle the limitations in previous search space design in terms of  the block counts and widths. The novelly designed routing blocks are utilized to construct the search space. The proposed chained cost estimation algorithm aims at optimizing both accuracy and model cost. The effectiveness of DenseNAS is demonstrated on both MobileNetV2- and ResNet- based search spaces. We leave more applications, \eg semantic segmentation, face detection, pose estimation, and more network-based search space implementations, \eg MobileNetV3~\cite{howard2019searching}, ShuffleNet~\cite{zhang2017shufflenet} and VarGNet~\cite{zhang2019vargnet}, for future work.

\section*{Acknowledgement}
This work was supported by National Key R\&D Program of China (No. 2018YFB1402600), National Natural Science Foundation of China (NSFC) (No. 61876212, No. 61733007 and No. 61572207), and HUST-Horizon Computer Vision Research Center. We thank Liangchen Song, Kangjian Peng and Yingqing Rao for the discussion and assistance.

{\small
\bibliographystyle{ieee_fullname}
\bibliography{egbib}
}

\appendix
\newcommand{\spaceline}{\specialrule{0em}{1pt}{1pt}}
\newcommand{\basicblock}[2]{$ 
    \left[
        \begin{tabular}{c}
        3 $\times$ 3, #1 \\ 
        3 $\times$ 3, #1
        \end{tabular} 
    \right] \times #2
$}

\newcommand{\bottleneck}[3]{$ 
    \left[
    \begin{tabular}{c} 
    1 $\times$ 1, #1 \\ 
    3 $\times$ 3, #1 \\
    1 $\times$ 1, #2
    \end{tabular}
    \right] \times #3
$}

\section{Appendix}
\subsection{Implementation Details}
Before the search process, we build a lookup table for every operation latency of the super network as described in Sec.~\textcolor{red}{3.3}. We set the input shape as $(3, 224, 224)$ with the batch size of $32$ and measure each operation latency on one TITAN-XP GPU. All models and experiments are implemented using PyTorch~\cite{paszke2017automatic}. 

For the search process, we randomly choose $100$ classes from the original 1K-class ImageNet training set. We sample $20\%$ data of each class from the above subset as the validation set. The original validation set of ImageNet is only used for evaluating our final searched architecture. The search process takes 150 epochs in total. We first train the operation weights for 50 epochs on the divided training set. For the last 100 epochs, the updating of architecture parameters ($\alpha, \beta$) and operation weights ($w$) alternates in each epoch. We use the standard GoogleNet~\cite{DBLP:conf/cvpr/SzegedyLJSRAEVR15} data augmentation for the training data preprocessing. We set the batch size to $352$ on $4$ Tesla V100 GPUs. The SGD optimizer is used with $0.9$ momentum and $4 \times 10^{-5}$ weight decay to update the operation weights. The learning rate decays from $0.2$ to $1 \times 10^{-4}$ with the cosine annealing schedule~\cite{DBLP:conf/iclr/LoshchilovH17}. We use the Adam optimizer~\cite{DBLP:conf/iclr/2015} with $10^{-3}$ weight decay, $\beta = (0.5, 0.999)$ and a fixed learning rate of $3 \times 10^{-4}$ to update the architecture parameters.

For retraining the final derived architecture, we use the same data augmentation strategy as the search process on the whole ImageNet dataset. We train the model for $240$ epochs with a batch size of $1024$ on $8$ TITAN-XP GPUs. The optimizer is SGD with $0.9$ momentum and $4 \times 10^{-5}$ weight decay.  The learning rate decays from 0.5 to $1 \times 10^{-4}$ with the cosine annealing schedule.

\begin{table*}[t]    
\centering
\captionof{table}{Architectures searched by DenseNAS in the ResNet-based search space.}
\begin{threeparttable}
    \begin{tabular}{c|c|c|c|c}
    \hline
    \textbf{Stage} & \textbf{Output Size} & \textbf{DenseNAS-R1} & \textbf{DenseNAS-R2} & \textbf{DenseNAS-R3} \\
    \hline
    1 & 112 $\times$ 112 & \multicolumn{3}{c}{3$\times$3, 32, stride 2}\\
    \hline

    2 & 56 $\times$ 56 & \basicblock{64}{1} & \basicblock{48}{1} & \begin{tabular}{c}
        \spaceline
        \bottleneck{48}{192}{1} \\
        \spaceline
    \end{tabular} \\
    \hline

    3 & 28 $\times$ 28 & \basicblock{72}{2} & \basicblock{72}{4} & \begin{tabular}{c}
        \spaceline
        \bottleneck{72}{288}{4} \\
        \spaceline
    \end{tabular} \\
    \hline

    4 & 14 $\times$ 14 & \begin{tabular}{c}
        \basicblock{176}{6} \\
        \spaceline
        \basicblock{192}{3}
    \end{tabular} & \begin{tabular}{c}
        \basicblock{176}{16} \\
        \spaceline
        \basicblock{208}{4}
    \end{tabular} & \begin{tabular}{c}
        \spaceline
        \bottleneck{176}{704}{16} \\
        \spaceline
        \bottleneck{208}{832}{4} \\
        \spaceline
    \end{tabular} \\
    \hline

    5 & 7 $\times$ 7 & \begin{tabular}{c}
        \basicblock{288}{1} \\
        \spaceline
        \basicblock{512}{1}
    \end{tabular} & \begin{tabular}{c}
        \basicblock{288}{2} \\
        \spaceline
        \basicblock{512}{1}
    \end{tabular} & \begin{tabular}{c}
        \spaceline
        \bottleneck{288}{1152}{2} \\
        \spaceline
        \bottleneck{512}{2048}{1} \\
        \spaceline
    \end{tabular} \\
    \hline

    6 & 1 $\times$ 1 & \multicolumn{3}{c}{average pooling, 1000-d fc, softmax} \\
    \hline
    \end{tabular}
    \end{threeparttable}
\label{tab: res_arch}
\end{table*}

\subsection{Viterbi Algorithm for Block Deriving}
The Viterbi Algorithm~\cite{forney1973viterbi} is widely used in dynamic programming which targets at finding the most likely path between hidden states. In DenseNAS, only a part of routing blocks in the super network are retained to construct the final architecture. As described in Sec.~\textcolor{red}{3.4}, we implement the Viterbi algorithm to derive the final sequence of blocks. We treat the routing block in the super network as each hidden state in the Viterbi algorithm. The path probability $p_{ij}$ serves as the transition probability from routing block $B_i$ to $B_j$. The total algorithm is described in Algo.~\ref{algo: viterbi}. The derived block sequence holds the maximum transition probability. 
    
\begin{algorithm}[htbp]
    \caption{The Viterbi algorithm used for deriving the block sequence of the final architecture.}
    \label{algo: viterbi}
    \KwIn{input block $B_0$, routing blocks $\{B_1, \dots, B_N\}$, ending block $B_{N+1}$, connection numbers $\{M_1, \dots, M_{N+1}\}$, path probabilities $\{p_{ji} | i=1, \dots, N+1, j=i-1, \dots, i-M_i\}$}
    \KwOut{the derived block sequence $X$}

    $P[0] \gets 1$ \tcp*[l]{record the probabilities}
    $S[0] \gets 0$ \tcp*[l]{record the block indices}
    \For{$i \gets 1, \dots, N+1$}{
        $P[i] \gets \displaystyle\max_{i-1 \le j \le i-M_i}(P[i-1] \cdot p_{ji})$\;
        $S[i] \gets \displaystyle\argmax_{i-1 \le j \le i-M_i}(P[i-1] \cdot p_{ji})$\;
    }
    $X[0] \gets B_{N+1}$\;
    $idx \gets N+1$\;
    $count \gets 1$\;
    \Do{$idx \neq 0$}{
        $X[count] \gets B_{S[idx]}$\;
        $idx \gets S[idx]$\;
        $count \gets count + 1$\;
    }
    $revers X$\;
\end{algorithm}

\subsection{Dropping-path Search Strategy}
The super network includes all the possible architectures defined in the search space. To decrease the memory consumption and accelerate the search process, we adopt the dropping-path search strategy~\cite{Understanding,cai2018proxylessnas} (which is mentioned in Sec.~\textcolor{red}{3.4}). When training the weights of operations, we sample one path of the candidate operations according to the architecture weight distribution $\{w_o^\ell | o \in \mathcal{O}\}$ in every basic layer. The dropping-path strategy not only accelerates the search but also weakens the coupling effect between operation weights shared by different sub-architectures in the search space. To update the architecture parameters, we sample two operations in each basic layer according to the architecture weight distribution. To keep the architecture weights of the unsampled operations unchanged, we compute a re-balancing bias to adjust the sampled and newly updated parameters.
\begin{equation}
    \mathtt{bias}_s = \ln \frac{\sum_{o \in \mathcal{O}_s} \exp(\alpha_o^\ell)}{\sum_{o \in \mathcal{O}_s} \exp({\alpha'}_o^\ell)},
\end{equation}
where $\mathcal{O}_s$ refers to the set of sampled operations, $\alpha_o^\ell$ denotes the original value of the sampled architecture parameter in layer $\ell$ and ${\alpha'}_o^\ell$ denotes the updated value of the architecture parameter. The computed bias is finally added to the updated architecture parameters.

\subsection{Implementation Details of ResNet Search}
We design the ResNet-based search space as follows. As enlarging the kernel size of the ResNet block causes a huge computation cost increase, the candidate operations in the basic layer only include the basic block~\cite{he2016deep} and the skip connection. That means we aim at width and depth search for ResNet networks. During the search, the batch size is set as 512 on 4 Tesla V100 GPUs. The search process takes 70 epochs in total and we start to update the architecture parameters from epoch 10. We set all the other search settings and hyper-parameters the same as that in the MobileNetV2~\cite{sandler2018mobilenetv2} search. For the architecture retraining, the same training settings and hyper-parameters are used as that for architectures searched in the MobileNetV2-based search space. The architectures searched by DenseNAS are shown in Tab.~\ref{tab: res_arch}.

\begin{table}[htbp]
    \centering
    \caption{Comparisons of different cost estimation methods.}
    \begin{threeparttable}
    \resizebox{1\columnwidth}{!}{
        \begin{tabular}{lcccc}
        \toprule
        \textbf{Estimation Method} & \textbf{FLOPs} & \textbf{GPU Latency} & \textbf{Top-1 Acc(\%)}\\
        \midrule
        local cost estimation & 396M & 27.5ms & 74.8 \\
        chained cost estimation & 361M & 17.9ms & 75.3 \\
        \bottomrule
        \end{tabular}}
        \end{threeparttable}
    \vspace{-15pt}
    \label{tab: cost-est}
\end{table}

\subsection{Experimental Comparison of Cost Estimation Method}
We study the design of the model cost estimation algorithm in Sec.~\textcolor{red}{4.5}. $1,500$ models are derived based on the randomly generated architecture parameters. Cost values predicted by our proposed chained cost estimation algorithm demonstrate a stronger correlation with the real values and more accurate prediction results than the compared local cost estimation strategy. We further perform the same search process as DenseNAS on the MobileNetV2~\cite{sandler2018mobilenetv2}-based search space with the local estimation strategy and show the searched results in Tab.~\ref{tab: cost-est}. DenseNAS with the chained cost estimation algorithm shows a higher accuracy with lower latency and fewer FLOPs. It proves the effectiveness of the chained cost estimation algorithm on achieving a good trade-off between accuracy and model cost.

\end{document}